# Fine-tuning vision foundation model for crack segmentation in civil infrastructures

K. Ge[1], C. Wang[2], Y. T. Guo[1*], Y. S. Tang[1], Z. Z. Hu[1], H. B. Chen[3]

[1]*Shenzhen International Graduate School, Tsinghua University, Shenzhen, China*
[2]*Department of Civil Engineering, Tsinghua University, Beijing, China*
[3]*School of Civil and Resource Engineering, University of Science and Technology Beijing, Beijing, China*

**Abstract**

Large-scale foundation models have become the mainstream deep learning method, while in civil engineering, the scale of AI models is strictly limited. In this work, a vision foundation model is introduced for crack segmentation. Two parameter-efficient fine-tuning methods, adapter and low-rank adaptation, are adopted to fine-tune the foundation model in semantic segmentation: the Segment Anything Model (SAM). The fine-tuned CrackSAM shows excellent performance on different scenes and materials. To test the zero-shot performance of the proposed method, two unique datasets related to road and exterior wall cracks are collected, annotated and open-sourced, for a total of 810 images. Comparative experiments are conducted with twelve mature semantic segmentation models. On datasets with artificial noise and previously unseen datasets, the performance of CrackSAM far exceeds that of all state-of-the-art models. CrackSAM exhibits remarkable superiority, particularly under challenging conditions such as dim lighting, shadows, road markings, construction joints, and other interference factors. These cross-scenario results demonstrate the outstanding zero-shot capability of foundation models and provide new ideas for developing vision models in civil engineering.

*\* Corresponding author:* Y. T. Guo (guoyutao@sz.tsinghua.edu.cn)

*Emails:* K. Ge (gk22@mails.tsinghua.edu.cn),

# 1. Introduction

Cracks are common in engineering structures, which may reduce the load-bearing capacity and stiffness of structures and lead to corrosion of internal reinforcements, reducing durability and causing structural failure [1]. Therefore, identifying cracks are important in structural health monitoring (SHM).

Traditionally, crack detection is carried out manually, which is costly, subjective and inefficient. Emerging SHM methods have paved the way for more automated, efficient and intelligent monitoring. Multiple nondestructive monitoring methods, including contact-based technologies such as sensors [2] and contactless methods such as ultrasound [3], infrared thermography [4], and unmanned aerial vehicles (UAVs) [5] are widely used in crack analyses.

The aim of crack segmentation is to classify crack images pixelwise to distinguish between cracks and backgrounds. More than a decade ago, the main approaches for crack segmentation were filters [6], wavelet transforms [7], and other image processing methods for denoising crack images.

In recent years, deep learning technologies have achieved rapid progress and are widely employed in computer vision (CV) tasks. Therefore, neural networks have become the mainstream approach for crack segmentation problems [8]. The models can be divided into two categories: CNN-based networks and Transformer-based [9] networks. The former can be seen as a series of stacked local filters, enhancing the model's receptive field through multiscale feature fusion. The latter effectively addresses the challenge of capturing long-distance dependencies through an attention mechanism.

A problem that cannot be ignored still exists: the crack segmentation model trained on a certain dataset may not be generalizable to other datasets. Pretrained models typically contain biases in the training set and tend to overfit on them but will perform poorly on unseen datasets. Chen noted the problem of cross-scenario/scale generalizability of defect detection models, where a pretrained crack segmentation model is not readily generalizable to sophisticated defect types or large-scale images, with the intersection over union (IoU) decreasing from 46.9% to 14.2 and 1.3%, respectively [10].

Crack segmentation is a highly class-imbalanced classification problem and is affected by many interference factors [11], including too-fine cracks, low resolution, different shooting distances, blurring, shadows, occlusions, traffic and pedestrian flows, and different working conditions. However, the crack images commonly used for training are much cleaner. These factors seriously hinder the deployment of pretrained models for identifying cracks in practical engineering.

The large-scale foundation models are bringing changes to various industries. Due to the complex architecture and extensive pretraining, these models have stronger performance compared to traditional small models. Inspired by the excellent zero-shot performance of foundation models, the purpose of this study is to introduce the powerful vision foundation model, explore their performance in crack segmentation, address the pain points of poor generalization and robustness of traditional architectures, and achieve intelligent crack segmentation empowered by large models.

In this paper, parameter-efficient fine-tuning (PEFT) technologies are applied to introduce the Segment Anything Model into crack segmentation. Two datasets with severe interference under different working conditions are collected. Evaluation of the proposed CrackSAM is focused on the inference performance on datasets with artificial noise and zero-shot performance on previously unseen datasets. Most previous studies have focused on analyzing high-quality images without considering scenarios with inadequate lighting, low resolution, and other non-ideal conditions. Furthermore, there is limited research on the zero-shot performance of pretrained crack segmentation models on unseen datasets, namely their generalization across datasets. However, this is precisely the most concerning engineering aspect. Without this, models will remain at the research stage and cannot be practically implemented. To the best of the authors' knowledge, this is the first work to fine-tune a large vision foundation model through PEFT technology for crack segmentation.

The advantage of CrackSAM lies in its powerful zero-shot capability, demonstrated by its robustness to noise and superior cross-dataset generalization compared to traditional classic architectures. However, its drawback is the high computational requirements and hardware demands during deployment. Especially in practical engineering applications, crack identification may need to rely on edge devices with limited computational power. Therefore, two deployment strategies are proposed: cloud computing and knowledge distillation-based edge computing.

The workflow of this study is shown in Figure 1. Note that the field test section has been supplemented in the appendix.

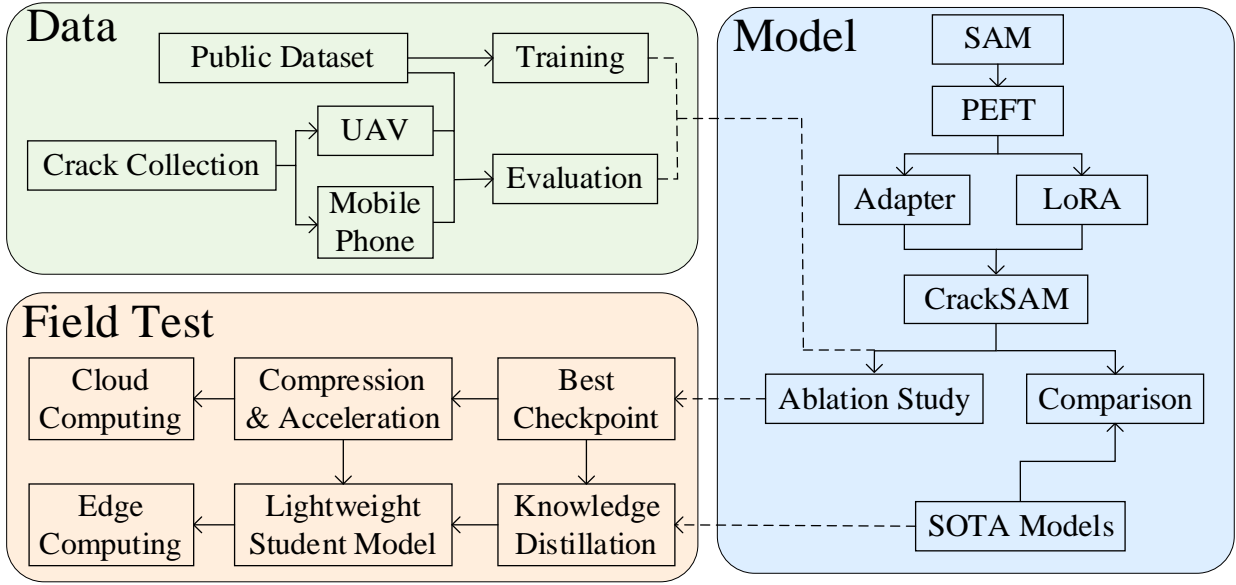

Figure 1 Workflow of this study.

The contributions of this paper are summarized as follows:
- Two PEFT methods, adapter [12] and low-rank adaptation (LoRA) [13], are employed to apply SAM for crack segmentation.
- Compared with twelve state-of-the-art (SOTA) models, the fine-tuned CrackSAM exhibits outstanding performance on datasets with artificial noise and on previously unseen datasets.
- Field test is conducted and the knowledge distillation-based lightweight strategy empowered by foundation models is proposed.

The remainder of this paper is arranged as follows. Section 2 reviews the relevant related studies. Section 3 describes the preparation of the datasets. Section 4 introduces the model architecture and fine-tuning methods. Section 5 presents the ablation studies and experimental results. Section 6 compares the proposed method with twelve other SOTA methods.

## 2. Relevant work

### 2.1 Models for crack segmentation

Semantic segmentation tasks emphasize global context information. Consequently, for CNN-based crack segmentation architectures, the UNet architecture [14] with skip connections and pyramid architecture fused with multiscale feature maps, such as FPN [15], PSPNet [16], and DeepLabV3+ [17], have achieved excellent performance in various crack segmentation tasks. Ren et al. [18] employed methods such as dilated convolutions, spatial pyramid pooling, and skip connections, to achieve feature aggregation and resolution reconstruction in crack segmentation. Dais et al. [19] combined UNet and FPN and integrated multiple backbones, such as VGG, ResNet, and MobileNet, to conduct comparative experiments on crack segmentation for masonry structures.

Due to the dominance of the Transformer architecture in recent years in CV tasks, Vision Transformer (ViT) [20], Swin Transformer [21], SegFormer [22] and many Transformer-based architectures have been widely used in crack analysis [23]. Shamsabadi et al. [24] used a TransUNet model with a hybrid CNN-ViT backbone to segment cracks. The performance of the proposed method was superior to that of CNN-based UNet and DeepLabV3+ while also exhibiting greater noise robustness. Guo et al. [25] employed a Swin Transformer architecture, achieving superior results on road surface cracks compared to models with UNet and ResNet as backbones.

However, small models trained on limited datasets still face issues of insufficient identification capabilities and poor generalizability.

## 2.2 Vision foundation models

Foundation models can be regarded as a new general paradigm of AI and have stronger intelligence than traditional models. These methods are usually based on Transformer architectures with billions of parameters. The realization of the foundation model entails large-scale pretraining on large datasets using massive GPUs. In CV, SAM [26] is a recently proposed foundation model for semantic segmentation, which is trained on more than 1 billion masks from 11 million images. Large-scale pretraining endows SAM with the ability to respond to various downstream tasks in a zero-shot manner. DINOv2 [27] is a vision foundation model trained in a self-supervised manner. Its backbone can be employed in various downstream tasks, such as image classification, instance recognition, semantic segmentation, and depth estimation.

However, directly applying vision foundation models for crack segmentation is not feasible. SAM tends to provide masks of all the distinguishable instances in the image [28], which are not useful for crack analysis. Quantitative analysis of cracks only requires a binary mask of the crack (Figure 2 (c)). Additionally, since SAM has not encountered semantics related to cracks, it also does not perform well in recognizing fine cracks (Figure 2 (b)). Consequently, it is necessary to fine-tune SAM to learn the specific semantics of cracks.

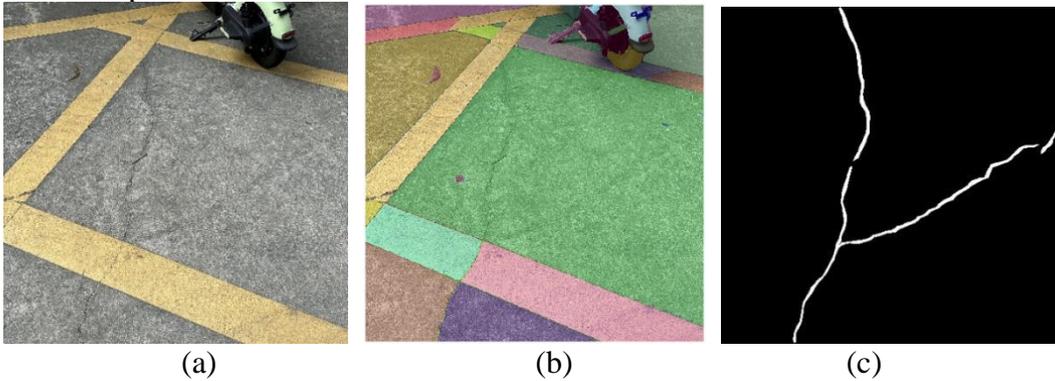

(a)  (b)  (c)

*Figure 2 Example images of segmentation results. (a) Input image. (b) Direct application of the SAM for crack images. (c) Ideal segmentation results.*

## 2.3 Transfer learning and zero-shot learning

Transfer learning is a technique that transfers the learned representations, features, and patterns acquired during the training process of a source task to a target task with limited labelled data.

Transfer learning has been widely practised in crack segmentation tasks. Zhou et al. [29] utilized the pretrained weights from Imagenet-1k to initialize the weights of the backbone to solve the data dependency of the Swin Transformer. The parameters of the backbone are frozen for the first 50 epochs and unfrozen for the latter 50 epochs. Gao et al. [30] established a hierarchical transfer learning architecture in which the pretrained model for localization and segmentation tasks inherited the trained backbone used for classification tasks.

Zero-shot learning is a subfield of transfer learning, which is used to classify test instances as an unseen class [31]. In crack segmentation tasks, there may be confusing interference from previously unseen objects in practical engineering situations, and the pretrained model needs to successfully classify this type of semantic information. Therefore, in this work, zero-shot learning can be evidenced by the ability to identify cracks in complex working conditions beyond the training set.

In AI-aided SHM, traditional transfer learning methods can be summarized as initializing the backbone of the model with high-quality pretrained weights (Figure 3(a)). During training, it is common in some studies to fine-tune only downstream networks (Figure 3(b)) or to freeze certain layers initially and then unfreeze and train all layers together. This full fine-tuning process requires significant GPU memory usage and sufficient data support.

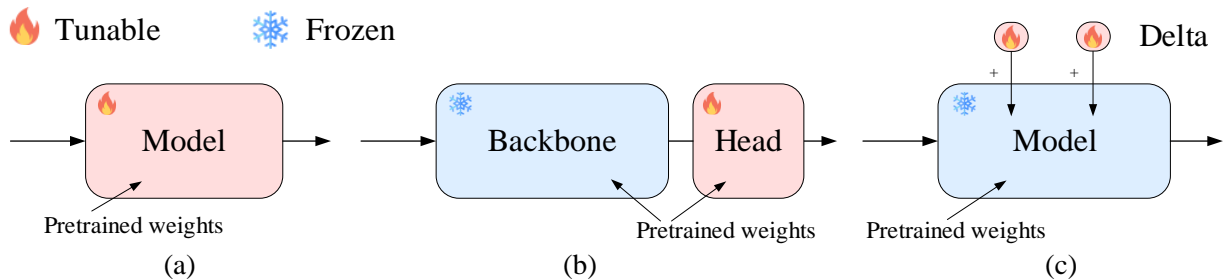

*Figure 3 Fine-tuning methods for pretrained models. (a) Full fine-tuning. (b) Fine-tuning of downstream networks. (c) PEFT.*

However, the scarcity of annotated datasets and computational resources make full fine-tuning of a foundation model in civil engineering impossible. Therefore, PEFT methods must be adopted.

## 2.4 PEFT

The PEFT is a lighter but more efficient fine-tuning method that is specifically designed for Transformer architectures. With fewer resources and training iterations, PEFT can preserve the original knowledge of pretrained foundation models and avoid catastrophic forgetting.

By introducing a few trainable parameters (less than 5%) that do not exist in the original network, the pretrained foundation model can be adapted for downstream tasks. Compared to other methods that train some or all the layers, PEFT essentially involves "delta-tuning" [32], as shown in Figure 3(c). For example, prefix-tuning achieves comparable performance in the full data setting by adding a trainable "soft prompt" to all the key and value matrices in Transformer layers [33] and even outperforms full fine-tuning in low data settings.

The PEEF of the SAM has already been applied to medical segmentation fields. Chen et al. [34] proposed the SAM-Adapter, which is effective in camouflaged object detection, shadow detection, and polyp segmentation. Wu et al. [35] also fine-tuned SAM with an adapter-based strategy on 19 medical image segmentation tasks, including CT, MRI, ultrasound images, fundus images, and dermoscopic images. Zhang et al. [36] applied the LoRA-based strategy to fine-tune the SAM for multiorgan segmentation in the Synapse dataset, which is on par with the SOTA method.

However, similar methods have not yet been applied in crack segmentation tasks. The most widely utilized PEFT technologies, adapter and LoRA, will be utilized in this work.

## 3. Data preparation

The design of dataset in this work is as follows: First, collect a large public dataset for fine-tuning, divide it into training, validation, and test set. Then, add artificial noise to the test set to simulate two potential scenarios: insufficient lighting and low-resolution. In addition, collect real-world datasets full of interference to test the generalization of the fine-tuned model on these challenging unseen datasets. The model is fine-tuned on the training set and inferred in a zero-shot manner on the test set, the test set with artificial noise, and the unseen datasets without changing the model settings.

### 3.1 Dataset for fine-tuning

The large labelled crack segmentation dataset collected by Khanhha [37] is utilized in this study. The Khanhha dataset is a union of multiple open-source subdataset, including CRACK500 [38], GAPs384 [39], CFD [40], AEL [41], CrackTree200 [42], CrackForest [40], and DeepCrack [43]. There are 9,603 images for training and 1,695 images for testing, with a resolution of 448×448. The crack dataset comes from diverse sources, including road surfaces, pavements, walls, and bridges, with different materials such as asphalt, cement, and concrete. Their clarity, shooting angle, and the size of the subject being captured are also different. For input crack images, there is no need for additional manual feature engineering and it is not necessary to train a separate model for a specific type of crack or a specific dataset. The richness of sources alleviates the tendency of the model to overfit to a specific small dataset and allows the model trained on this large dataset to perceive cracks of different materials, structures, working conditions and scales, thus leading to generalization.

However, inconsistent annotation thicknesses in different datasets can have a certain negative impact on the model. The annotation of the subdataset CrackTree200 is too fine compared to that of the other datasets, with the width of the crack masks being only one pixel. This inconsistent labelling strategy made it difficult for previous studies [44][45] to identify cracks in the CrackTree200 dataset; therefore, this dataset was manually relabelled by experts.

## 3.2 Datasets collected for zero-shot

Two crack datasets are captured for evaluating the model's zero-shot performance: Road420 and Facade390. The pixel-level binary masks are obtained via expert annotation. The images and masks were converted into RGB and grayscale images and downsampled to a size of 448×448.

### 3.2.1 Road420

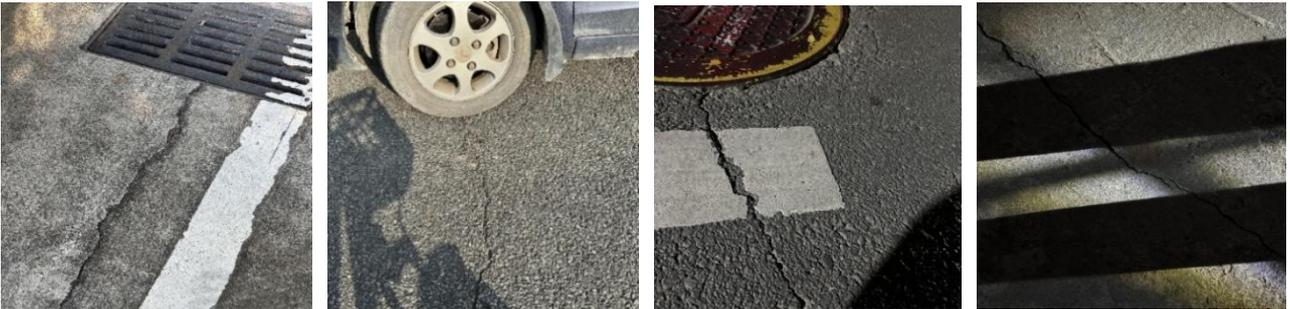

*Figure 4 Sample images of Road420.*

The Road420 consists of 420 images of asphalt concrete and cement concrete road surfaces with cracks. The pictures contain considerable interfering information, such as shadows, occlusions, road signs, vehicles, manhole covers, people, and leaves. Some small cracks are difficult for the naked eye to recognize after downsampling. Some of the images were taken at night. The image semantics of some interfering factors have never appeared in the Khanhha dataset, and the deliberate introduction of such interference makes zero-shot learning very challenging on this dataset. All the images were captured with an iPhone 14 Pro. Representative sample images are shown in Figure 4.

### 3.2.2 Facade390

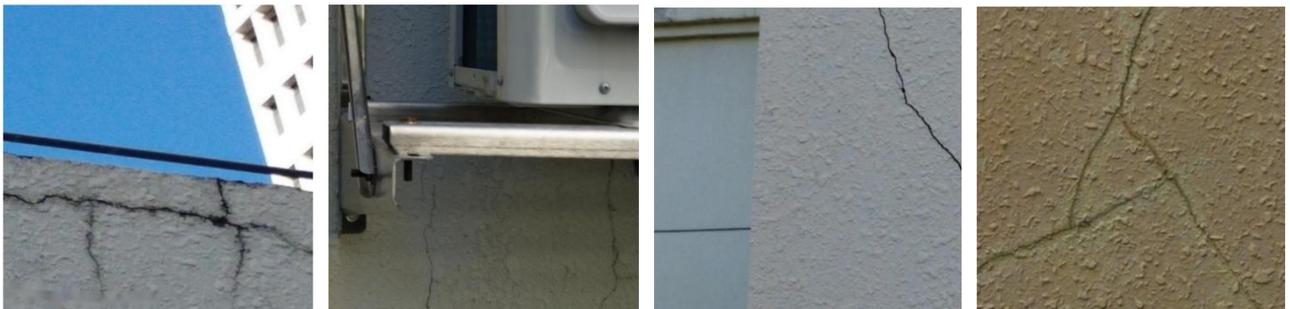

*Figure 5 Sample images of Facade390.*

The Facade390 is composed of cracks on the exterior walls and columns of buildings captured by UAV. Because the UAV must maintain a safe distance from the building during operation and may experience displacement during hovering, the captured images may be blurry, and some fine cracks may not be clearly visible. The identification of these cracks is susceptible to interference from various factors, such as wall stains, peeling, water traces, shadows, paint, vegetation, construction joints, and other extraneous factors. The cracks in Facade390 are generally not structural cracks but may introduce unsettling risks such as water seepage. The UAV employed in this study was a Dajiang Mini 3. Some representative sample images are presented in Figure 5.

*3.2.3 Concrete3k*

Concrete3k is a large, ready-made dataset with 3,000 image-label pairs of concrete cracks contributed by [44][46], it will also be leveraged for zero-shot images.

## 4. Methodology

## 4.1 Overall architecture

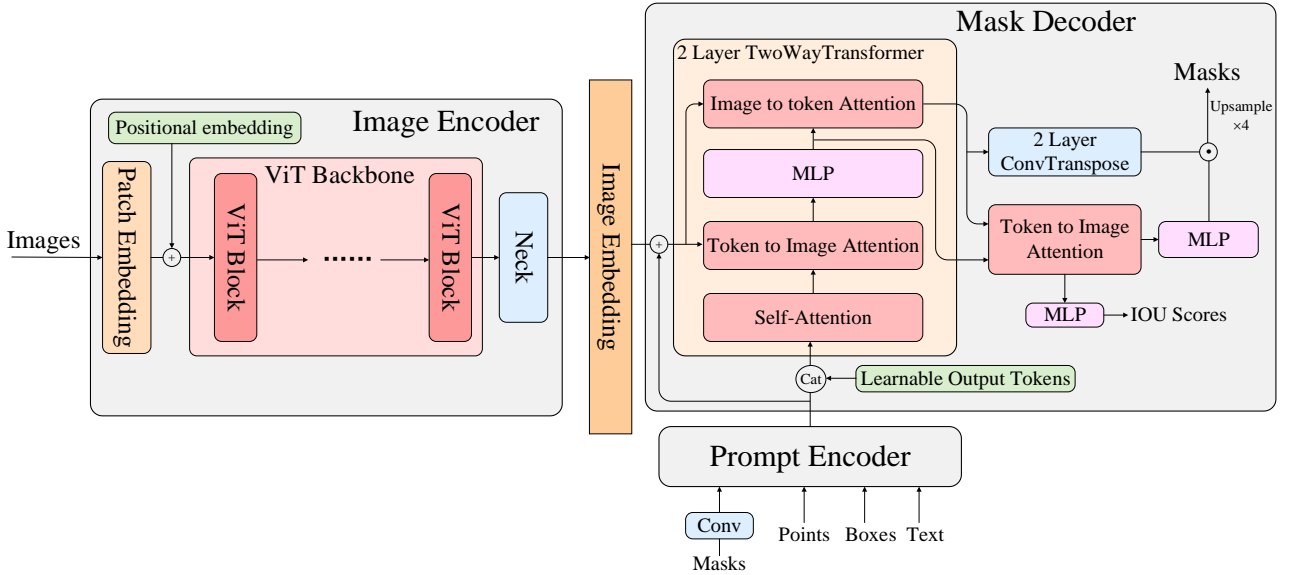

*Figure 6 The original overall architecture of the SAM.*

The SAM is composed of three key components: an image encoder, a prompt encoder and a mask decoder, as shown in Figure 6. The latter two parts are much lighter than the image encoder. Considering that previous works [34][35][36] fine-tuning the SAM mostly did not alter the SAM architecture itself, proving the maturity and reliability of its architecture, this study also refrains from modifying the overall architecture and instead chooses to introduce additional delta for fine-tuning.

*4.1.1 ViT block*

The ViT block is composed of two parts: window attention and a multilayer perceptron (MLP), as shown in Figure 7.

First, the input patches $x_p \in \mathbb{R}^{H \times W \times C}$ undergo a window partition with a size of *w* and are separated into *N* nonoverlapping windows $x \in \mathbb{R}^{N \times w \times w \times C}$, where $N = HW/w^2$. The window size is set to 14. Then, a multihead self-attention approach is applied to *x*.

*x* is divided along the channel and feed into multiple attention heads. In the *i*-th head, query vector *Q* and key-value pairs *K* and *V* are obtained via a learnable linear layer:

$$Q_i / K_i / V_i = W_{Q_i/K_i/V_i} x_i + b_{Q_i/K_i/V_i} \tag{1}$$

Dot production is computed to calculate the similarity scores between *Q* and *K*, which are divided by the square root of the dimension size *K* for scaling. The result is normalized by a softmax activation function after a learnable positional embedding is added to it. The results are multiplied by *V* to obtain the output for each head:

$$Atten_i = softmax(\frac{Q_i K_i^T}{\sqrt{d_k}} + pos)V_i \tag{2}$$

The outputs of each head are concatenated, and feed into a linear layer. Finally, the windows are reorganized into their original shapes $x_o \in \mathbb{R}^{H \times W \times C}$.

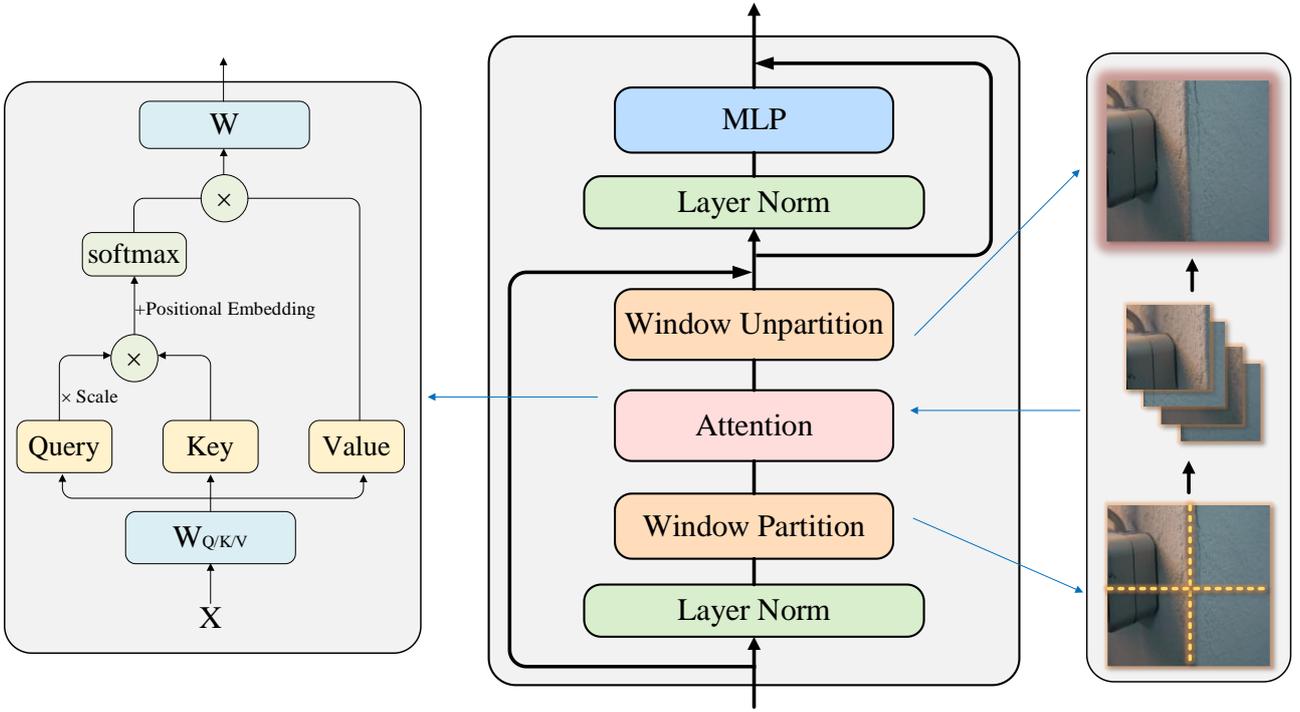

Figure 7 The architecture of the ViT block.

The MLP is a fully connected neural network stacked in multiple layers that expands the original dimensions fourfold and compresses it back with Gaussian error linear unit (GELU) activation [47].

The window attention and the MLP inputs are normalized through layer normalization [48]. Moreover, residual connections [49] are added to each module.

*4.1.2 Image encoder*

The image encoder is an MAE [50] pretrained ViT and consists of a patch embedding layer, a learnable positional embedding, a ViT backbone and a neck. The patch embedding layer is a 16×16 convolution with a stride of 16. The absolute positional embedding is added to each position of the feature map. The backbone is a stack of ViT blocks. There are three options based on different size: ViT-H, ViT-L, and ViT-B. The embedding dimension, number of blocks and number of attention heads for ViT-B are 768, 12 and 12, respectively. For ViT-L, it is 1024, 24 and 16. For ViT-H, it is 1280, 32 and 16. In the neck, the image embedding is fed to a pointwise convolution and a 3×3 convolution to reduce the dimension to 256. Each convolution is followed by layer normalization.

The output of the image encoder is a 16× downscaled image embedding with 256 dimensions.

*4.1.3 Prompt encoder*

The prompt encoder receives sparse or dense prompts. However, in this task, the segmentation object (crack) is determined, so the input of the prompt is simplified to none. The default embedding is a learnable vector that is added to each position of the image embedding.

*4.1.4 Mask decoder*

A set of learnable tokens is concatenated with the prompt embeddings. The obtained tokens, the image embedding, and its positional embedding, are fed into a 2-layer two-way Transformer.

For additional details about the two-way Transformer, refer to the original code [26]. Within the two-way Transformer, the following steps are performed: first, self-attention is conducted on the input tokens; second, cross-attention is applied from the tokens to the image embedding, where the tokens are regarded as queries and the image embedding is treated as keys and values; then, the tokens are updated through an MLP; and finally, an image-to-token cross-attention is carried out. After each attention and MLP layer, a residual connection and a layer normalization are added.

A two-layer transposed convolution with a stride of 2 and a kernel size of 2×2 upsamples the image embedding to 1/4-size. GELU activation is performed after convolution, and layer normalization is added between the layers.

Tokens are updated through the final cross-attention layer and a 3-layer MLP. The output is a linear classifier with a shape of (*num_class*, 32). The shape of the image embedding is (32, *H*/4, *W*/4). Finally, the low-resolution masks with a shape of (num_class, *H*/4, *W*/4) are obtained by a pointwise product. High-resolution masks can be derived through bilinear interpolation.

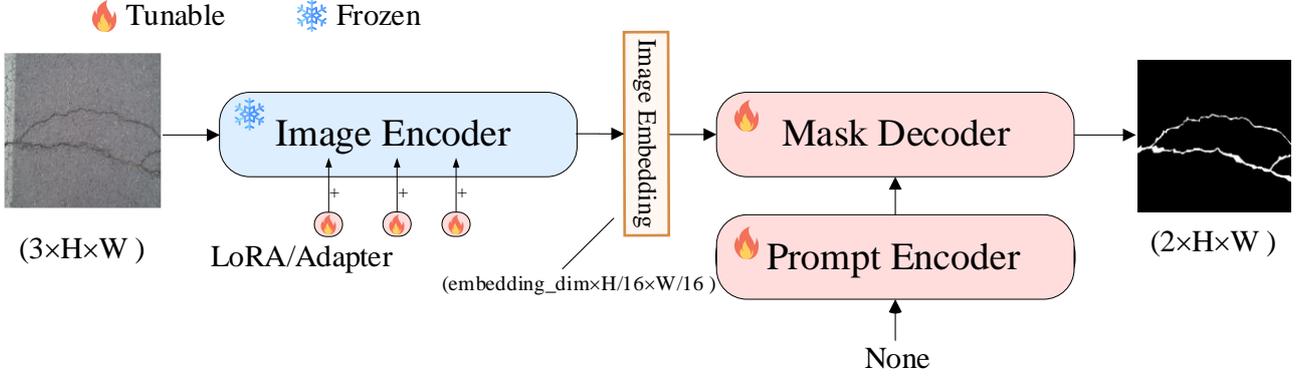

*Figure 8 Architecture of the proposed CrackSAM.*

Because the main parameters are concentrated in the ViT blocks, PEFTs are conducted on these blocks. The lightweight prompt encoder and mask decoder are also fine-tuned together [36]. The general architecture of the fine-tuned SAM, CrackSAM, is illustrated in Figure 8.

## 4.2 Adapter

Adapter is the most commonly used method in PEFT. There are many variants of adapters, which can be either sequential or parallel. In this work, an adapter is sequentially inserted behind the attention layer and parallelly inserted at the MLP [35], as shown in Figure 9. The GELU activation function is employed in the middle to increase nonlinearity.

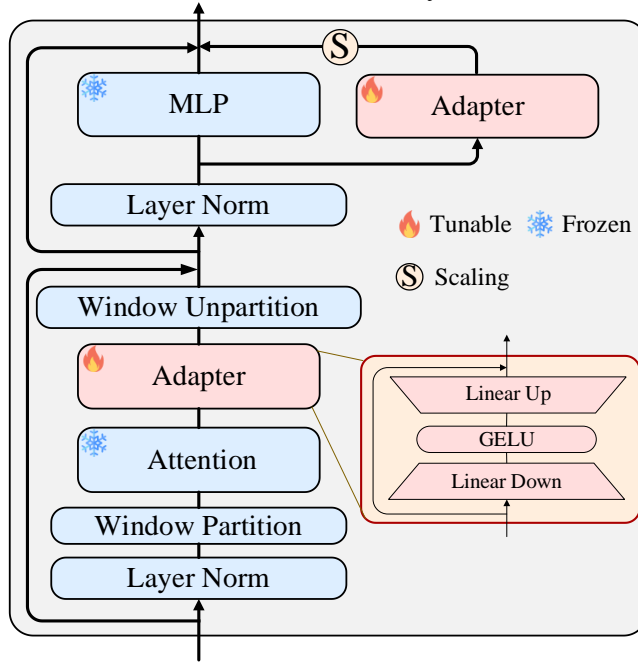

*Figure 9 Fine-tuning strategy for the adapter.*

Initially, the adapter employs a downprojection linear layer with parameters $W_{down} \in \mathbb{R}^{d \times m}$ to project the original *d*-dimensional features to a smaller dimension *m*. Subsequently, a nonlinear activation function is applied, followed by an upprojection layer with parameters $W_{up} \in \mathbb{R}^{m \times d}$ to restore the features to the *d*-dimensional space. A residual connection is incorporated in this process.

The middle dimension $m$ is constrained such that $m \ll d$. Denoting the input as $x$ and the output after adaptation as $x'$, the transformation is expressed as Eq. (3).

$$x' = (W_{up} \cdot \text{GELU}(W_{down}x + b_{down}) + b_{up}) + x \tag{3}$$

For a parallel adapter, a scaling factor $s$ is needed to control the extent of the update of the adapter. Given an input $x_m$ for the MLP and its adapted output $x_m'$, the formula is as follows in Eq. (4).

$$x_m' = s \cdot (W_{up} \cdot \text{GELU}(W_{down} \cdot \text{LN}(x_m) + b_{down}) + b_{up}) + \text{MLP}(\text{LN}(x_m)) + x_m \tag{4}$$

During fine-tuning, the weights of the attention layer and MLP are frozen, and only the weights of the adapter are trained. When the middle dimension is a multiple of the input dimension, the fine-tuning adapter at this point is equivalent to fine-tuning the newly added MLP.

### 4.3 Low-rank adaptation

LoRA is a reparameterization method that converts the original parameters in a neural network into a parameter-efficient form [32]. A neural network usually consists of many full-rank matrix operations. When migrating to downstream tasks, LoRA assumes that the pretrained model has a small intrinsic rank, and the updating of weights can be achieved on this small subspace.

For a pretrained weight matrix $W_o \in \mathbb{R}^{d \times k}$, a bypass $\Delta W \in \mathbb{R}^{d \times k}$ is added to constrain the update of its weight, and $\Delta W$ is decomposed into the product of matrices $A \in \mathbb{R}^{d \times r}$ and $B \in \mathbb{R}^{r \times k}$ using low-rank decomposition, with the rank $r \ll \min(d, k)$. As Figure 10 shows, for the original path $y = W_o x$, the updated result $y'$ is:

$$y' = (W_O + \Delta W)x = W_O x + ABx \tag{5}$$

During fine-tuning, the weight matrix $W_o$ is kept frozen, and only matrices $A$ and $B$ are fine-tuned. Matrix $A$ is initialized using random Gaussian initialization, and matrix $B$ is initialized to 0 [13].

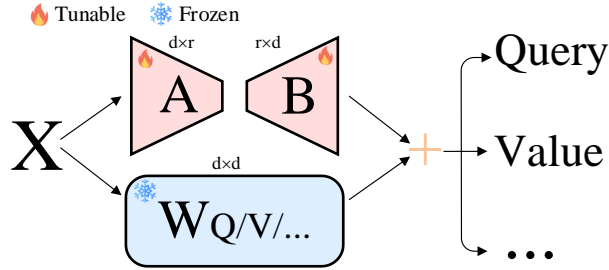

*Figure 10 Fine-tuning strategy for the LoRA.*

LoRA can theoretically be added to any set of weights. In this way, fine-tuning the LoRA is equivalent to full fine-tuning. From a parameter-efficient perspective, LoRA is always added to the attention weights, typically on the query and value components. Due to the low-rank decomposition, LoRA has fewer parameters than does the adapter.

## 5. Experiment and results

### 5.1 Implementation details

The primary loss function for semantic segmentation is cross-entropy (Eq. (6)). However, due to the highly imbalanced nature of crack segmentation, the cross-entropy loss tends to rapidly converge to zero during training, thereby allowing the background to dominate the loss [29]. Hence, a more effective approach involves using a weighted combination of cross-entropy and Dice loss (Eq. (7)), as demonstrated in Eq. (8).

$$L_{CE} = -y \log(\hat{y}) - (1-y)\log(1-\hat{y}) \tag{6}$$

$$L_{Dice} = 1 - \frac{2|X \cap Y|}{|X| + |Y|} \tag{7}$$

$$L = \lambda L_{CE} + (1-\lambda)L_{Dice} \tag{8}$$

In Eq. (6), *y* represents the ground truth labels, taking values of 0 or 1. $\hat{y}$ denotes the probability of the predicted labels. In Eq. (7), *X* corresponds to the mask regions of the true labels, and *Y* represents the mask regions of the predicted labels. The parameter *λ* in Eq. (8) is a weighting coefficient and is set to 0.2 here.

The learning rate is adjusted using a "poly" policy incorporating a warm-up strategy. For the first 300 iterations, the learning rate linearly increases from 0 to the initial learning rate of 0.0004. Then, the learning rate is dynamically scaled by multiplying by $(1 - \frac{iter - warm\_up}{max\_iter})^{power}$, with the power set to 6. The maximum iteration limit is set to 140 epochs, with a batch size of 8. The model optimization utilizes AdamW, with parameters $\beta_1$, $\beta_2$, and weight_decay set to 0.9, 0.999, and 0.01, respectively.

A threshold of 0.5 is chosen for mask binarization. Basic data augmentation techniques, including random rotation and random flipping, are employed. Pretrained weights of the SAM are loaded and frozen before training. The best-performing checkpoints for the delta and head in the validation set are saved and selected for subsequent testing. The model is established on the PyTorch framework and trained on only one 24 GB RTX3090 GPU.

## 5.2 Evaluation metrics

The precision (Pr), recall (Re), F1-score (F1), and IoU are employed to evaluate the segmentation performance of the model, as defined in Eq. (9) - Eq. (12).

$$\Pr = \frac{TP}{TP + FP} \tag{9}$$

$$\mathrm{Re} = \frac{TP}{TP + FN} \tag{10}$$

$$F1 = \frac{2 \cdot \Pr \cdot \mathrm{Re}}{\Pr + \mathrm{Re}} \tag{11}$$

$$IoU = \frac{TP}{TP + FP + FN} \tag{12}$$

where true positive (TP) denotes pixels representing cracks that are correctly classified, false positive (FP) represents background pixels erroneously classified as cracks, and false negative (FN) indicates crack pixels misclassified as background.

## 5.3 Ablation study

In this section, an ablation study is conducted on the parameter setting of the proposed architecture. For the adapter, it is necessary to study the size of the middle dimension and scaling factor. For the LoRA, the positions where the LoRA is applied and the size of the rank should be investigated. In addition, the impacts of the size of the backbone and the combination of two fine-tuning methods are studied.

The following experiment discusses the Pr, Re, F1, and IoU metrics (Eq. (9) - Eq. (12)) when a well-trained crack segmentation model is inferring on the test set. Moreover, the model's generalization ability is evaluated by directly applying the fine-tuned model to three new datasets (Road420, Facade390, and Concrete3k) in a zero-shot manner without any additional training and measuring the IoU metric.

### 5.3.1 CrackSAM_adapter

As Table 1 shows, introducing a few parameters is sufficient to achieve excellent transfer to downstream tasks. Even when the middle dimension is set to 1, the model can still achieve decent precision. Increasing the middle dimension continuously improved the metrics on the test set, but may lead to a decrease in generalizability. When the middle dimension is 32, the IoU on the test set decreases by only 0.3% compared to when the middle dimension is 64. However, there are improvements of 1.9%, 3.7%, and 4.8% when performing zero-shot learning on Road420, Facade390, and Concrete3k, respectively.

*Table 1 Ablation study on the middle dimension of the adapter.*

| Middle dimension | Metric of inference on test set | | | | IoU of zero-shot on the new dataset | | |
|---|---|---|---|---|---|---|---|
| | Pr | Re | F1 | IoU | Road420 | Facade390 | Concrete3k |
| dim=1 | 0.7554 | 0.7786 | 0.7515 | 0.6270 | 0.5310 | 0.4618 | 0.6743 |
| dim=16 | 0.7664 | 0.7968 | 0.7696 | 0.6479 | 0.6139 | 0.4772 | 0.6461 |
| dim=32 | 0.7676 | 0.7965 | 0.7704 | 0.6495 | 0.6149 | 0.4718 | 0.6718 |
| dim=64 | 0.7674 | 0.8002 | 0.7719 | 0.6513 | 0.6033 | 0.4548 | 0.6412 |

Note that the model's IoU on the Facade390 dataset is relatively low. This is because Facade390 is composed mainly of cracks in building exterior wall materials, which are very fine compared to road cracks. The masks in the training set are mostly coarse segment-wise annotations, resulting in a lower IoU during the zero-shot operation on Facade390. In fact, a well-tuned CrackSAM model can accurately detect the majority of cracks in Facade390 (Re > 0.9). Given this, in ablation experiments, priority is given to evaluating the generalization ability of the proposed method based on the Road420 and Concrete3k datasets.

The scaling factor *s* is introduced to balance the task-agnostic features generated by the frozen backbone and the task-specific features generated by the tunable parallel adapters. As shown in Table 2, setting the scaling factor to 0.2 can yield better performance in terms of generalization.

*Table 2 Ablation study on the scaling factor of the adapter.*

| Scaling factor | Metric of inference on test set | | | | IoU of zero-shot on new dataset | | |
|---|---|---|---|---|---|---|---|
| | Pr | Re | F1 | IoU | Road420 | Facade390 | Concrete3k |
| *s*=0.1 | 0.7671 | 0.7953 | 0.7693 | 0.648 | 0.6042 | 0.4487 | 0.6597 |
| *s*=0.2 | 0.7676 | 0.7965 | 0.7704 | 0.6495 | 0.6149 | 0.4718 | 0.6718 |
| *s*=0.5 | 0.7716 | 0.7934 | 0.7706 | 0.6499 | 0.609 | 0.4354 | 0.6635 |
| *s*=1 | 0.7702 | 0.7958 | 0.7709 | 0.6500 | 0.6006 | 0.4313 | 0.6426 |
| *s*=2 | 0.7751 | 0.7902 | 0.7707 | 0.6494 | 0.5981 | 0.4586 | 0.6548 |

*5.3.2 CrackSAM_LoRA*

*Table 3 Ablation study on the rank of the LoRA.*

| Rank | Metric of inference on test set | | | | IoU of zero-shot on new dataset | | |
|---|---|---|---|---|---|---|---|
| | Pr | Re | F1 | IoU | Road420 | Facade390 | Concrete3k |
| *r=1* | 0.7509 | 0.7941 | 0.7585 | 0.6352 | 0.6176 | 0.4494 | 0.6516 |
| *r=4* | 0.7620 | 0.7918 | 0.7639 | 0.6416 | 0.6222 | 0.4544 | 0.6798 |
| *r=8* | 0.7656 | 0.7925 | 0.7665 | 0.6448 | 0.6201 | 0.4601 | 0.6800 |
| *r=16* | 0.7657 | 0.7947 | 0.7687 | 0.6473 | 0.6200 | 0.4573 | 0.6727 |

According to Table 3, similar to an adapter, fine-tuning a LoRA with a rank set to 1 is quite effective, and at this point, the parameters of the LoRA component are only 0.16 M. As the rank increases, the metrics on the test set continuously improve. The model's generalization reaches saturation when the rank is set to 4 or 8. The decrease in generalization caused by over-parameterization for the LoRA is similar to that for the adapter.

The LoRA layer can be applied to the query, key, value, and output matrices in the attention layer. As shown in Table 3 and Table 4, when the rank is 8 and the LoRA layer is applied to only the query, even though the number of parameters is equivalent to the situation when the rank is 4 and the LoRA is applied to both the query and value, the latter achieves higher metrics. Applying LoRA to all four matrices has a similar effect on excessively increasing the rank, resulting in a slight improvement in the metrics of the test set but a decrease in the zero-shot capability.

Table 4 Ablation study on the weight type of the LoRA.

| Weight type | Metric of inference on test set | | | | IoU of zero-shot on the new dataset | | |
|---|---|---|---|---|---|---|---|
| | Pr | Re | F1 | IoU | Road420 | Facade390 | Concrete3k |
| $W_q$ | 0.7489 | 0.7964 | 0.7575 | 0.6344 | 0.5800 | 0.5122 | 0.6562 |
| $W_q$, $W_v$ | 0.7656 | 0.7925 | 0.7665 | 0.6448 | 0.6201 | 0.4601 | 0.6800 |
| $W_q$, $W_k$, $W_v$, $W_o$ | 0.7717 | 0.7887 | 0.7690 | 0.6476 | 0.6183 | 0.4602 | 0.6501 |

When comparing adapter and LoRA, the former demonstrates slightly higher metrics on the test set, while the latter exhibits better generalization performance.

*5.3.3 Combine two PEFT methods or use neither*

Here, comparisons are made between simultaneously using both PEFT methods and not using both methods, i.e., employing only the traditional fine-tuning of the head (Figure 3(b)).

Three different parameter scales of the adapter and LoRA combinations are tested. According to Table 5 and Table 4, the performance improvement brought by the combination of multiple PEFT methods is not significant. Considering the obvious increase in the computational cost, there is no significant advantage to combine multiple PEFT methods.

When fine-tuning only the prompt encoder and mask decoder, the performance significantly decreases. This clearly demonstrates the superiority of the PEFT algorithm over traditional fine-tuning methods because completely freezing the backbone makes it more challenging for the model to extract semantic information related to cracks.

Table 5 Experimental results of combining two methods and using neither method.

| Delta type | Metric of inference on test set | | | | IoU of zero-shot on the new dataset | | |
|---|---|---|---|---|---|---|---|
| | Pr | Re | F1 | IoU | Road420 | Facade390 | Concrete3k |
| No PEFT, fine-tune only head | 0.6951 | 0.7188 | 0.6843 | 0.5564 | 0.4826 | 0.4288 | 0.6059 |
| adapter($s^*$=0.2, dim$^*$=8) + LoRA(qv$^*$, r$^*$=2) | 0.7596 | 0.7959 | 0.7657 | 0.6438 | 0.6132 | 0.4560 | 0.6628 |
| adapter(s=0.2, dim=16) + LoRA(qv, r=4) | 0.7637 | 0.8005 | 0.7703 | 0.6488 | 0.6188 | 0.4639 | 0.6798 |
| adapter(s=0.2, dim=32) + LoRA(qv, r=8) | 0.7664 | 0.7959 | 0.7696 | 0.6485 | 0.6230 | 0.4862 | 0.6835 |

Note: [*1]s = scaling factor; [*2]dim = middle dimension; [*3]qv = apply LoRA to query and value matrices; [*4]r = rank.

*5.3.4 Backbone size*

According to Table 6, the size of the backbone has a significant impact on the model. The larger the backbone is, the more powerful the segmentation ability after fine-tuning. Progressing from ViT-B to ViT-L and then to ViT-H, the segmentation performance and generalization ability improved both for the adapter and the LoRA. This observation aligns with the scaling law of large language models. This difference may be attributed to the richer features extracted by the stronger backbone and the smaller intrinsic dimension. With the same parameter configuration, fine-tuning becomes more effective for large-scale backbones. Therefore, this paper regards only ViT-H as the backbone for CrackSAM.

Table 6 Ablation study on the size of the backbone.

| Delta type | Backbone | Metric of inference on test set | | | | IoU of zero-shot on the new dataset | | |
|---|---|---|---|---|---|---|---|---|
| | | Pr | Re | F1 | IoU | Road420 | Facade390 | Concrete3k |
| adapter | ViT-B | 0.7574 | 0.7920 | 0.7610 | 0.6379 | 0.5859 | 0.4680 | 0.6573 |
| adapter | ViT-L | 0.7611 | 0.8004 | 0.7682 | 0.6464 | 0.6263 | 0.4700 | 0.6672 |
| adapter | ViT-H | 0.7676 | 0.7965 | 0.7704 | 0.6495 | 0.6149 | 0.4718 | 0.6718 |
| LoRA | ViT-B | 0.7512 | 0.7823 | 0.7523 | 0.6286 | 0.5905 | 0.4787 | 0.6557 |
| LoRA | ViT-L | 0.7623 | 0.7849 | 0.7608 | 0.6379 | 0.6162 | 0.4862 | 0.6791 |
| LoRA | ViT-H | 0.7620 | 0.7918 | 0.7639 | 0.6416 | 0.6222 | 0.4544 | 0.6798 |

# 6. Comparison with SOTA models

Semantic segmentation models typically consist of three main components: the backbone, neck, and head. The backbone plays a crucial role in extracting high-level and semantically rich features. The neck assists in fusing information across multiple scales, while the head can be viewed as a decoder, obtaining the desired mask through aggregation and upsampling.

Twelve other models that have performed well in semantic segmentation are selected for comparative experiments with CrackSAM. These include VGG-UNet [37], Swin-UPerNet [21], MobileNet [51], UNet-FCN and UNet-PSPNet [14], ResNet-DeepLabV3+ [49], ViT-UPerNet [20], SegFormer [22], HRNet-FCN [52], and ResNet-PSPNet [53]. The selected models have significant differences in terms of parameter quantity and include various types of backbones, such as CNN-based UNet, ResNet-50, and ResNet-101, and attention-based architectures such as Swin-T, ViT-B, and Mix Transformer (MiT-B5).

The architectures for comparative models are configured using the settings from the OpenMMLab segmentation toolkit [55]. The main training configuration is closely aligned with the CrackSAM. The parameters are initialized using the pretrained weights from OpenMMLab.

This study evaluates the performance of CrackSAM from two perspectives: robustness and generalization.

## 6.1 Evaluation on datasets with artificial noise

Artificial noise is introduced into the test set to assess the model robustness. This paper primarily investigates the following two cases when introducing artificial noise:

Case 1: This situation represents images with insufficient lighting in practical engineering. For the input image $I$, reduce its brightness and apply Gaussian blur:

$$I' = (I - bri) * K \tag{13}$$

*Bri* represents the brightness, $*$ denotes the convolution operation, and $K$ represents the Gaussian kernel. Specifically, the JPG image is converted to the HSV colour space. Then, 50 is subtracted from the V channel to achieve a decrease in brightness. Next, a 2D Gaussian filter with a kernel size of 9x9 is used to smooth the JPG image.

Case 2: This situation represents blurry images with insufficient resolution in practical engineering. Apply serious blur to the input image, followed by downsampling:

$$I' = (I * K) \downarrow_S \tag{14}$$

$\downarrow$ denotes the downsampling operation, and $S$ is the scaling factor. Gaussian blur is applied to the image with a kernel size of 21x21, after which the image is downsampled to half its original size via cubic interpolation, followed by interpolation back to the original size.

The experimental results are listed in Table 7, and some predicted masks are shown in Figure 11. Figure 11 (a)-(d) are from case 1, simulating a dim environment. The remaining figures are from case 2, indicating a fuzzy situation. As the figure shows, the proposed CrackSAM performs well in identifying various forms of cracks, such as linear (Figure 11 (c)), branched (Figure 11 (d)), and webbed (Figure 11 (b)). It is capable of predicting different materials (asphalt, concrete), various structures (road surfaces, walls, etc.), and diverse crack thicknesses, brightness levels, and contrast ratios.

Figure 11 (a) is a noncrack image. Neither CrackSAM_adapter nor CrackSAM_LoRA provides any masks, while the other four comparative models identify construction joints and paint as cracks.

For extremely dark conditions (Figure 11 (c)), the CrackSAM can still identify some cracks, although a few may be indistinguishable from the background. In extremely blurry situations (Figure 11 (d), (e), (h)), CrackSAM can find most cracks that are difficult for the naked eye to discern, while other models can hardly detect the cracks in the image.

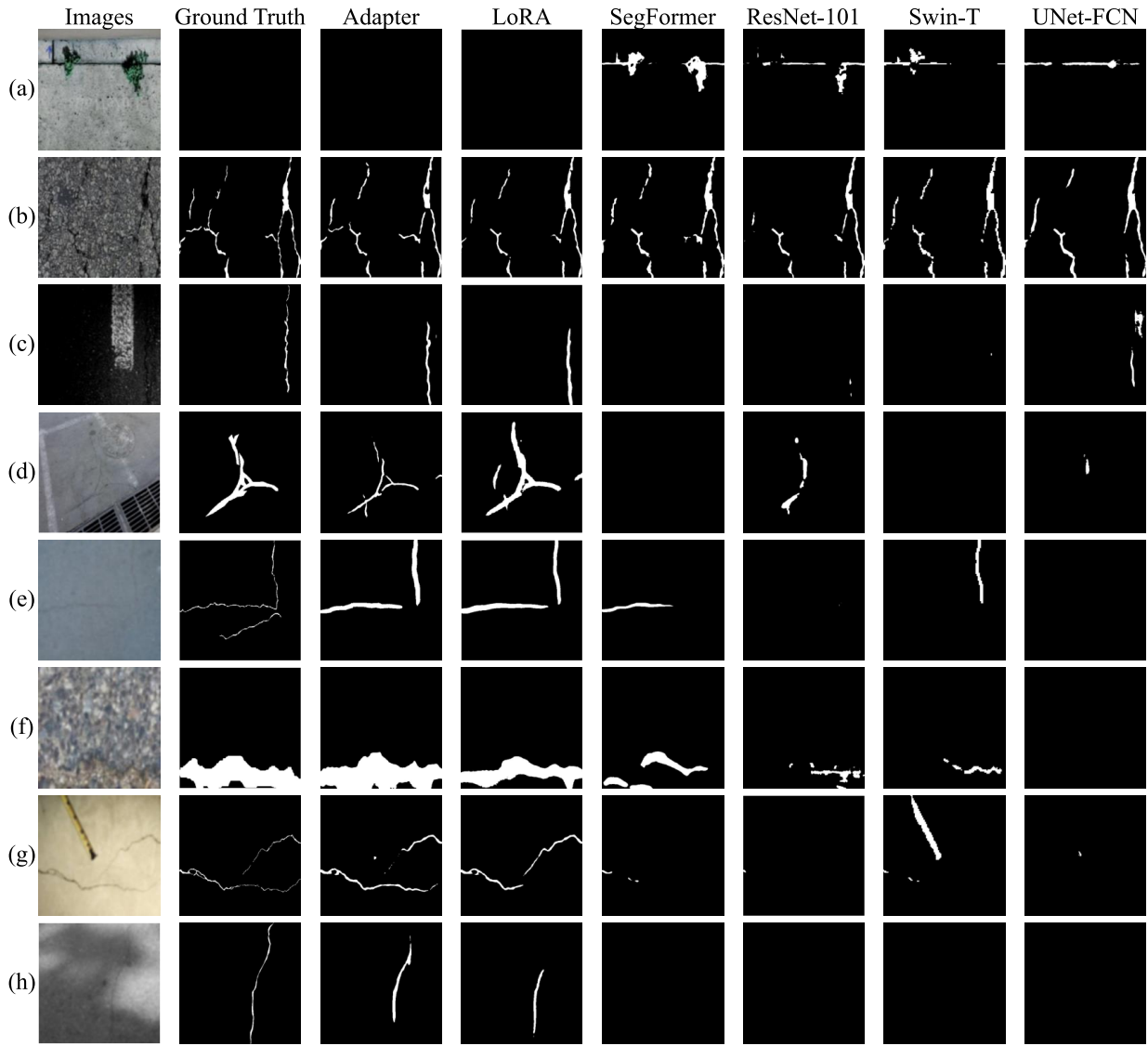

*Figure 11 Inference results of comparative experiments on the test set with artificial noise.*

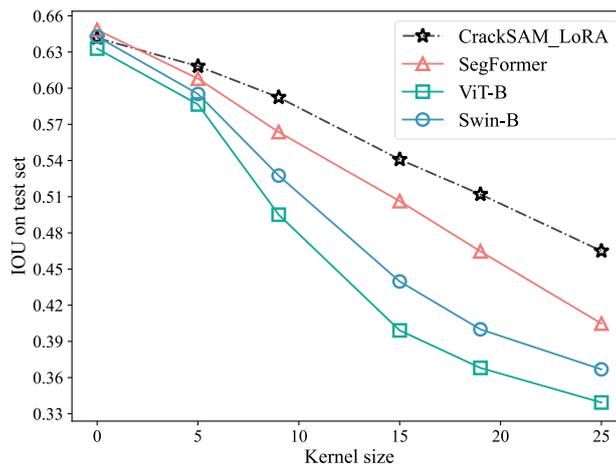

*Figure 12 Variation in the IoU for different models under different Gaussian kernel sizes.*

Figure 12 studied the IoU of different models with different Gaussian kernel sizes when adding a Gaussian blur to images in the test set. As shown in the figure, CrackSAM demonstrates much greater robustness than do the other models. Despite the comparable performances of the compared models on the unprocessed test set, their IoUs significantly decrease when a blur is added.

Table 7 reveals that almost all the models achieved satisfactory results on the original test set

(IoU ≥ 0.62). Excluding UNet-PSPNet, the maximum gap between CrackSAM and the other 11 SOTA models is 4.6% on the original test set. However, significant differences emerge on the noisy test set, with variations reaching 27.0% and 42.0% in two specific cases. The classic UNet architecture performs poorly on severely blurred test sets, with UNet-FCN and UNet-PSPNet experiencing accuracy drops of 54.4% and 60.3%, respectively, after adding severe blur, while CrackSAM_LoRA experiences only a 23.4% drop. Among all the models, CrackSAM_adapter is the most accurate model on the test set, while CrackSAM_LoRA is the most accurate on the noisy test set. This shows that the robustness of the CrackSAM is much better than that of traditional models.

## 6.2 Zero-shot performance on unseen datasets

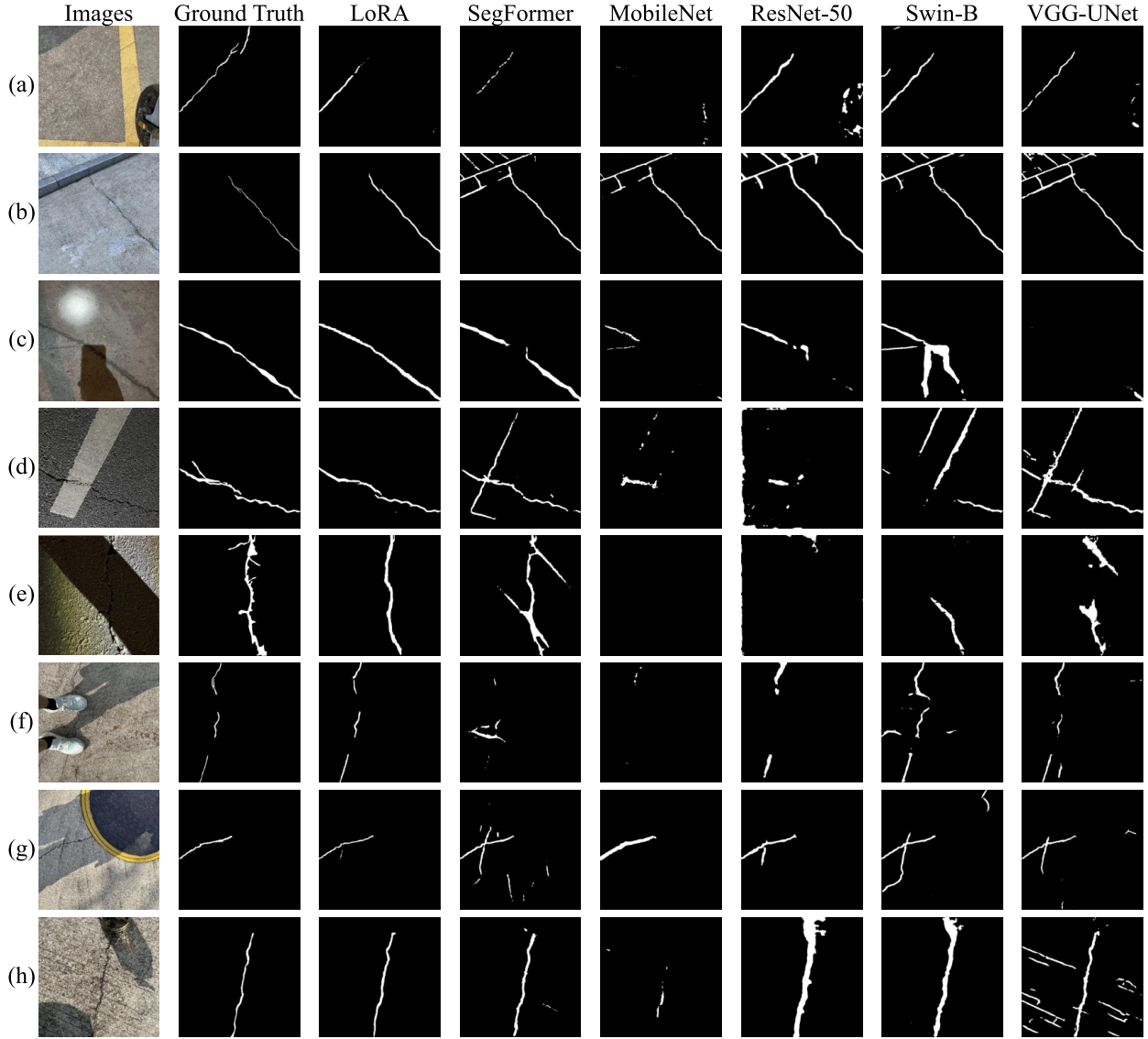

*Figure 13 Zero-shot results of comparative experiments on Road420.*

As Figure 13 shows, the proposed model yields satisfactory predictions at various scales and under different interference conditions. In Figure 13 (a), cracks captured from a distant view become challenging to discern by the naked eye after downsampling, yet the AI models used in this experiment can still identify cracks. For cracks captured at night, as Figure 13 (c), (d), and (e) show, CrackSAM maintains highly accurate predictions, particularly in Figure 13 (e), where the cracks almost merge with shadows, while CrackSAM still accurately identifies the cracks. However, other models are strongly affected by shadows and road markings, which can result in numerous artifacts. In Figure 13 (b), the CrackSAM correctly distinguishes between construction joints and cracks, but other models are misled by the sidewalk. In Figure 13 (f), CrackSAM is not affected by occlusion and accurately outputs three segments of cracks, whereas other models either produce artifacts or fail to recognize the three complete segments. CrackSAM correctly segments the cracks in Figure 13 (g) with little influence from tire tracks on the road surface. In Figure 13 (h), the segmentation performance of the CrackSAM remains unaffected by the presence of a cup and refracted light.

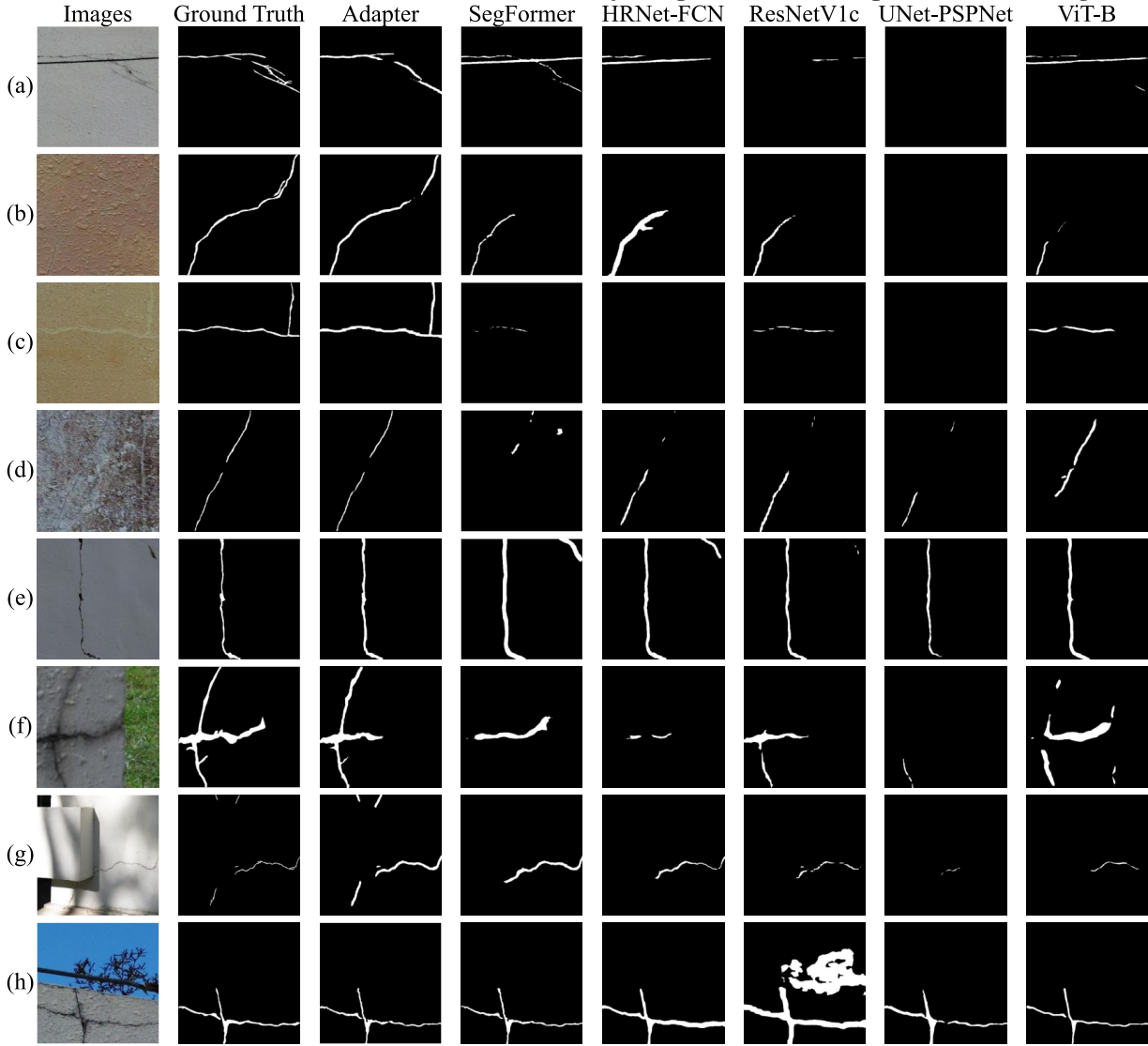

*Figure 14 Zero-shot results of comparative experiments on Facade390.*

Due to the impact of annotation thickness in Facade390, which generally results in a lower IoU, it is necessary to combine the figures of the segmentation results when conducting comparative analysis on this dataset. As depicted in Figure 14, the proposed model can effectively identify cracks in an automated and efficient manner when combined with UAVs, even when images are captured from different angles and distances. In Figure 14 (a), other models struggle to distinguish between construction joints and cracks, whereas CrackSAM can. Figure 14 (b) and (c) showcase red building facades with peeling, where the CrackSAM successfully segments small cracks, while the other models fail to do so. Figure 14 (d) shows walls with paint and peeling, where due to severe interference, comparative models cannot segment complete cracks, whereas CrackSAM provides results closest to the ground truth. Figure 14 (f) shows surface cracks on the column with damp stains and grass; the segmentation mask of CrackSAM closely resembles the actual crack morphology. In Figure 14 (g), at the junction of the beam and column with tree shadows, the CrackSAM identifies all four cracks, including the two small cracks at the top of the image.

According to Table 7, similar to robustness, the generalization gap among different models is also substantial, reaching 42.9%, 33.0%, and 31.1% on three new datasets. Both fine-tuned versions of CrackSAM exhibit excellent cross-dataset generalization capabilities. CrackSAM_LoRA is the best-performing model during the zero-shot operation on the Road420 and Concrete3k datasets, while CrackSAM_adapter performs better on Facade390. Among the twelve SOTA models, SegFormer is the best in terms of both robustness and generalizability. However, the proposed CrackSAM achieves a significant improvement in IoU metrics compared to SegFormer on the two noisy test sets and three new datasets, with increases of up to 11.1%, 10.8%, 7.0%, 2.1%, and 4.1%, indicating a notable improvement in performance.

Table 7 also illustrates the significance of studying the robustness and generalizability of crack segmentation models, as they truly impact the feasibility of models for real-world applications. In summary, through comparative experiments, the proposed CrackSAM achieved the best results in terms of test set accuracy, robustness, and zero-shot performance.

The robust generalization ability of the CrackSAM is primarily attributed to the power of the large backbone and the effectiveness of the PEFT method. The deep ViT layers endow it with stronger feature extraction capabilities, which contribute to better performance in the presence of noise. Meanwhile, CrackSAM retains the high-quality weights of SAM during fine-tuning. This is akin to having encountered various data distributions previously, leading to better generalization when facing unseen dataset.

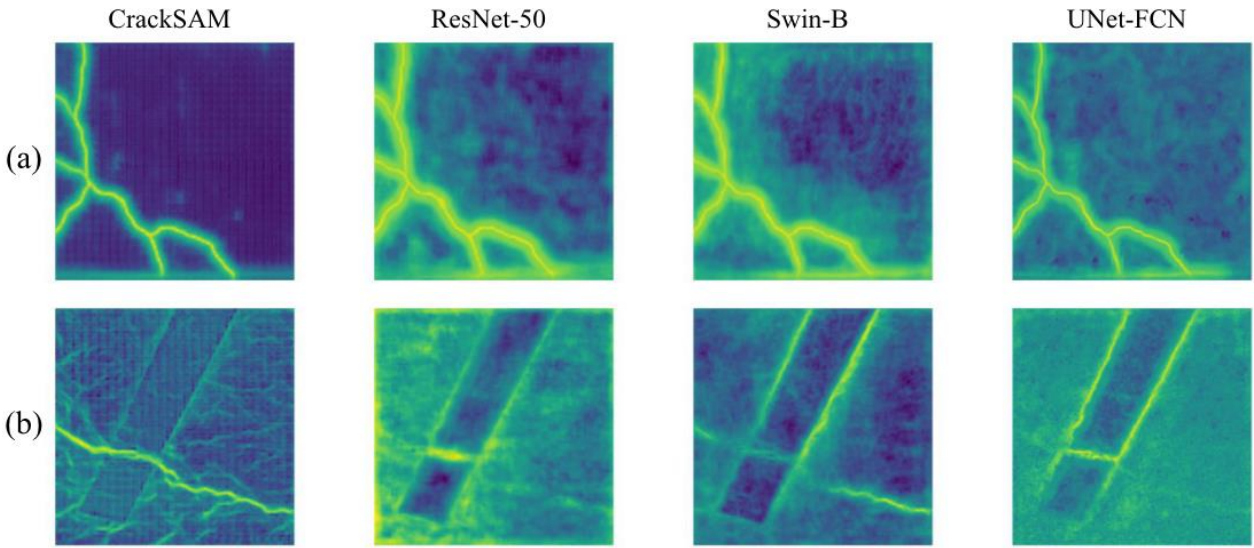

*Figure 15 Visualization of the channel with an index of 1 of the output logits from different models.*

Visualization of the channel with an index of 1 (corresponding to crack categories) of the output logits from different models are shown in Figure 15. In Figure 15 (a), it can be observed that all models have achieved relatively accurate prediction masks, but CrackSAM exhibits the highest discrimination between cracks and background in its logits, with the ratio of grayscale values corresponding to foreground and background being the highest. This high discrimination capability of CrackSAM results in better identification of cracks, especially in the presence of noise or other interference when compared to other models.

In Figure 15 (b), CrackSAM provides the most precise mask, while other models incorrectly classify road markings as cracks. From the feature maps, it is evident that CrackSAM encodes the saliency of cracks without focusing on road textures and markings, unlike other models that are influenced by road markings leading to misclassifications.

*Table 7 Comparison of the noisy dataset and unseen dataset with other SOTA models.*

| Model | Backbone | Parameters | Test set | Noisy test set 1 | Noisy test set 2 | Road420 | Facade390 | Concrete3k |
|---|---|---|---|---|---|---|---|---|
| | | | - | -50 bri[*] + blur(k[*]=9) | ×1/2 + blur(k=21) | Zero-shot | Zero-shot | Zero-shot |
| CrackSAM_adapter (dim=32, s=0.2) | ViT-H | 641.9 M (Tunable 9.1 M) | **0.6495** | 0.5466 | 0.4763 | 0.6149 | **0.4718** | 0.6718 |
| CrackSAM_LoRA (qv, rank=4) | ViT-H | 637.2 M (Tunable 4.4 M) | 0.6416 | **0.5782** | **0.4915** | **0.6222** | 0.4544 | **0.6798** |
| VGG-UNet | VGG16 | 53.91 M | 0.6419 | 0.4337 | 0.3472 | 0.5126 | 0.4547 | 0.5152 |
| Swin-UPerNet | Swin-T | 58.9 M | 0.6199 | 0.4745 | 0.3963 | 0.4628 | 0.4065 | 0.4778 |
| Swin-UPerNet | Swin-B | 120.0 M | 0.6428 | 0.5003 | 0.3857 | 0.5262 | 0.4593 | 0.5655 |
| MobileNet-V3 | MobileNet-V3 | 3.28 M | 0.6208 | 0.5068 | 0.3738 | 0.4447 | 0.4322 | 0.6154 |
| UNet-FCN | UNet | 28.99 M | 0.6255 | 0.4218 | 0.2852 | 0.4531 | 0.3677 | 0.4682 |
| UNet-PSPNet | UNet | 28.97 M | 0.5594 | 0.3535 | 0.2222 | 0.3555 | 0.3163 | 0.5262 |
| ResNet-DeepLabV3+ | ResNet-101 | 60.2 M | 0.6402 | 0.5115 | 0.4088 | 0.3827 | 0.4399 | 0.5791 |
| ResNet-DeepLabV3+ | ResNet-50 | 41.2 M | 0.6395 | 0.5077 | 0.3947 | 0.3918 | 0.4402 | 0.5601 |
| ResNet-PSPNet | ResNetV1c-101 | 65.59 M | 0.6346 | 0.5110 | 0.4207 | 0.4084 | 0.4327 | 0.5544 |
| ViT-UPerNet | ViT-B | 142.1 M | 0.6328 | 0.4714 | 0.3554 | 0.5171 | 0.4276 | 0.6027 |
| SegFormer | MiT-B5 | 82.0 M | 0.6484 | 0.5204 | 0.4436 | 0.5817 | 0.4622 | 0.6533 |
| HRNet-FCN | HRNet-W18 | 9.63 M | 0.6356 | 0.5055 | 0.4434 | 0.4172 | 0.4214 | 0.6322 |

Note: [*1]bri= brightness; [*2]k= kernel size.

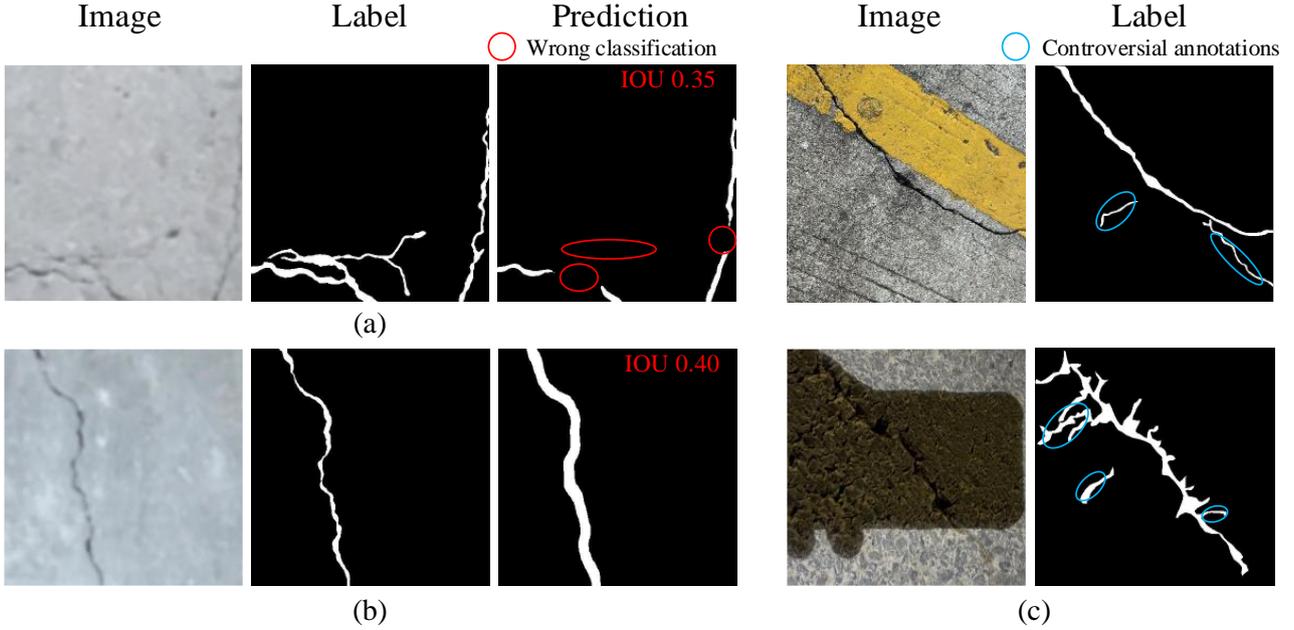

*Figure 16 Several prediction situations with low IoUs. (a) Wrong classification. (b) Thicker prediction mask. (c) Controversial annotations subject to subjective judgements.*

The issues affecting the model's accuracy can generally be categorized into three situations. In the first scenario, the model fails to recognize the semantics of a particular object, as illustrated in Figure 16 (a). This reflects a deficiency in the model. The second situation arises when the model correctly identifies a crack and successfully outputs its mask. However, because the output mask is much coarser than the annotation mask, the IoU is lower, while the recall is very high, as depicted in Figure 16 (b). This situation is considered acceptable. In the third type, annotators, when providing high-quality annotations on high-resolution images, may label not only the main crack but also adjacent minor cracks and defects. Whether these tiny defects should be annotated depends on the annotator's subjective judgement (Figure 16 (c)). After downsampling, information about these minor defects is severely lost, rendering them unidentifiable. This situation often occurs in the annotation of asphalt road cracks, where asphalt has complex textures and background noise, and asphalt and cracks have similar brightness and contrast ratios, leading to ambiguous situations and inaccurate segmentation results.

## 7. Conclusions

This paper fine-tuned the segment anything model using PEFT methods for crack segmentation. The proposed CrackSAM was fine-tuned on more than 11k images. Two new labelled datasets comprising 810 images were collected through smartphones and UAVs for zero-shot. The main conclusions of this paper are as follows:

(1) PEFT technology was utilized, with the image encoder of SAM frozen and a trainable delta (adapter and LoRA) introduced on the ViT backbone. Fine-tuning was applied to the head and delta with only one RTX 3090 GPU. The trainable parameters are approximately 1% of the original SAM. The pretrained vision foundation model can be effectively introduced into crack segmentation.

(2) The proposed CrackSAM based on PEFT improved the IoU score greatly compared to the traditional method of fine-tuning only the head. The fine-tuning of CrackSAM followed the scaling law, where using ViT-H as the backbone instead of ViT-B resulted in additional performance gains.

(3) Excessive over-parameterization can possibly enhance the performance on the test set but may not necessarily generalize to other datasets, with an increase in computational cost and a simultaneous decrease in generalization. Therefore, the fine-tuning design entails a tradeoff between complexity, performance, and generalization.

(4) CrackSAM worked exceptionally well in cross-scale and cross-scenario situations, exhibiting

strong robustness and generalizability. Satisfactory results were achieved on the test set by most SOTA models, but when faced with datasets with noise and unseen datasets with interference, CrackSAM significantly outperforms traditional models, demonstrating the powerful feature extraction and zero-shot capability of the foundation model.

The proposed CrackSAM is much larger than the existing SOTA models, but considering the various factors that may affect crack segmentation in real-world deployment, the introduction of a large foundation model is meaningful.

For situations where there is background noise in images, such as blurred images or asphalt road surfaces with rich texture, it is possible to introduce generative networks such as GAN, Diffusion, etc., to enhance the features of the image, achieving purposes like denoising or super-resolution.

The technique used in this paper can also be extended to segment other types of structural defects, such as earthquake damage segmentation, rebar corrosion segmentation, and structural water seepage identification, which enables "segment everything" in SHM. Moreover, the powerful backbone of SAM can be retained, while configuring it with other types of heads for various different downstream tasks such as defect classification and detection. These issues will be the focus of future work.

## 8. Declarations

### 8.1. Funding

The authors gratefully acknowledge the financial support provided by the National Natural Science Foundation of China, Grant No. 52308179, and by the Cross-disciplinary Research and Innovation Fund Research Plan of Tsinghua Shenzhen International Graduate School, Grant No. JC2023002.

### 8.2. Conflicts of interest

There are no conflicts of interest for this paper.

### 8.3. Data availability

Models, pretrained weights, and the labelled datasets will be publicly available on https://github.com/KG-TSI-Civil/CrackSAM after acceptance.

## Appendix A. Model deployment

First, it is necessary to compress and accelerate the well-trained CrackSAM. The ONNX model is exported in inference mode, while performing constant folding optimization. Then, OpenVINO framework is used to convert the ONNX file into IR (Intermediate Representation) format, while compressing the precision from Float32 to Float16. Next, the IR model is read and compiled on the server. The throughput is tested on the server using the benchmark_app tool provided by OpenVINO, with a throughput of 3.55 FPS (Frames Per Second) on two Intel Xeon Gold 6248R CPUs. API services are set up on cloud servers using frameworks such as Flask. Edge devices can send inference requests through API interfaces, and obtain masks from the responses.

Cloud computing can utilize the computing power of servers, but may be affected by poor network conditions. Edge computing can infer directly on edge devices, but is greatly limited by computing power. This study also explores edge computing solutions and conducts a field test.

The edge device used here is Raspberry Pi 4B. Compared with professional edge computing devices equipped with powerful GPUs, such as Jetson Nano, Raspberry Pi is cheaper and has much lower computing power, simulating scenarios where computing resource is very limited in engineering. The well-trained CrackSAM is used as the teacher model, and a lightweight model is chosen as the student model, which is trained through knowledge distillation (see in Appendix B). The student model selected here is HRNet-FCN.

The inference of Raspberry Pi is accelerated using the Intel Neural Compute Stick 2 (NCS2). After compressing and accelerating the student model, the inference speed on the server's CPU reaches 131.52 FPS, while on the Raspberry Pi with NCS2, the inference speed is 3.71 FPS.

A campus road is selected for field test. A small experimental car equipped with a Raspberry Pi

4B chip is controlled by a Python program. The IR file is loaded and compiled, and the video stream from the camera is read and processed for inference using MJPG-Streamer. The testing scenario is shown in Figure A.1.

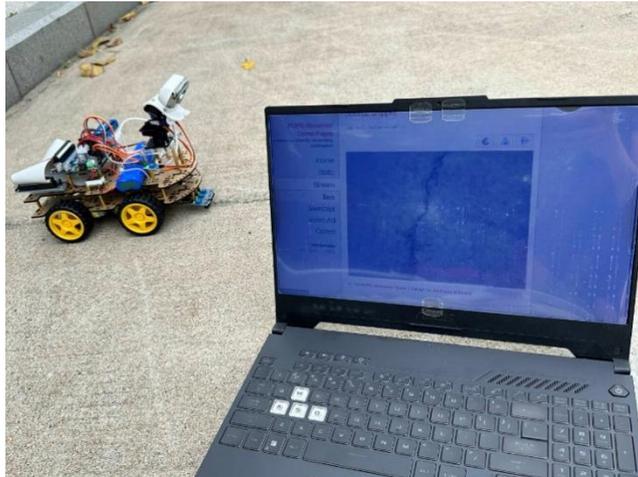

*Figure A.1 Field test.*

Figure A.2 shows some inference results of the saved video frames on edge device. Despite the low resolution, the obtained results are satisfactory.

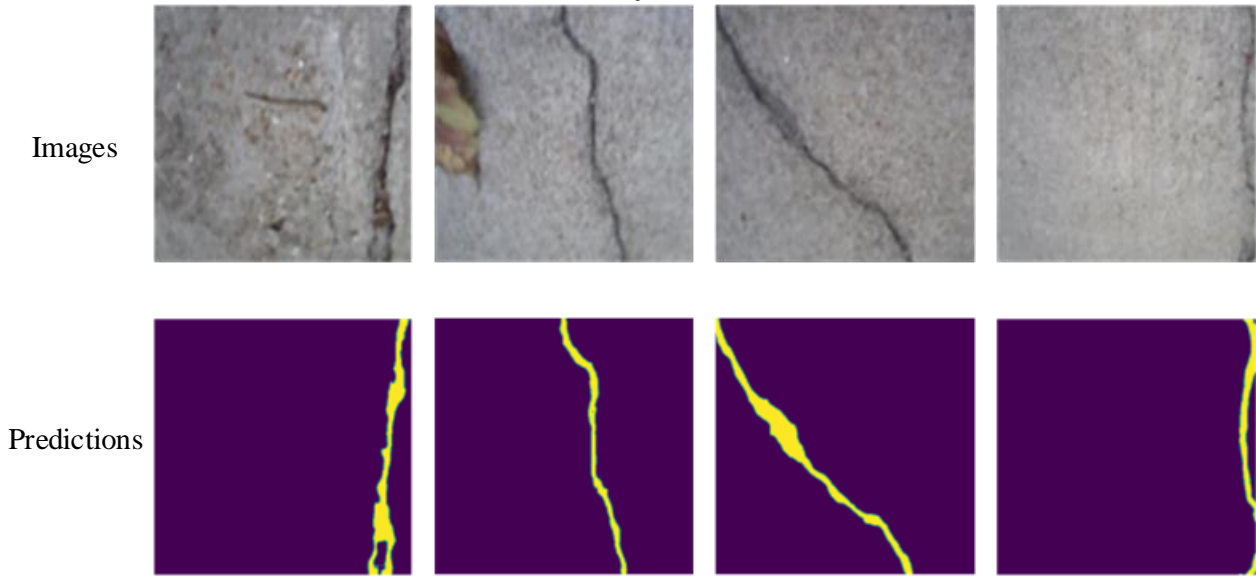

*Figure A.2 The crack images taken in the field test and the prediction masks obtained by edge computing.*

## Appendix B. Knowledge distillation

Knowledge distillation aims to enhance the performance of compact student models by leveraging supervision from heavy teacher models [56]. As shown in Figure B.1, compared to the traditional method of calculating hard loss $L_{hard}$ based on outputs and ground truth, knowledge distillation requires an additional soft loss $L_{soft}$ as a constraint.

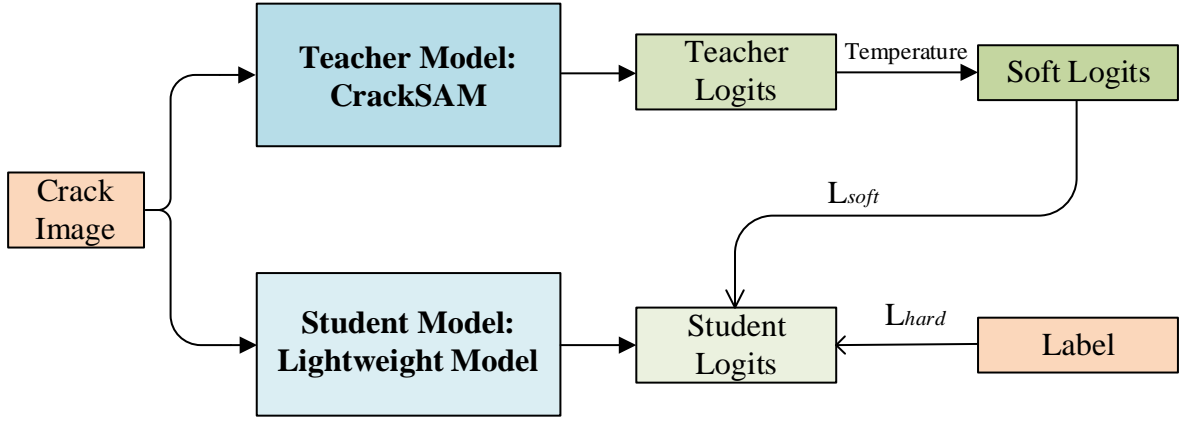

*Figure B.1 Schematic diagram of knowledge distillation for crack segmentation empowered by CrackSAM.*

The logits size of student model needs to be consistent with that of the teacher model. If not, sampling needs to be done through interpolation or convolution. This study adopts the method of channel-wise distillation [57]. Given a set of logits $I \in \mathbb{R}^{C \times H \times W}$, where $C$ is the number of channels and $H \times W$ is the size of the feature map, softmax activation with temperature $T$ is applied to each feature map along the channels, resulting in an activated map $AP \in \mathbb{R}^{C \times H \times W}$, as shown in Eq. (B.1):

$$AP_c = \frac{\exp(I_{c,i}/T)}{\sum_{i=1}^{W \cdot H} \exp(I_{c,i}/T)} \quad (B.1)$$

where $c=1, 2, ..., C$ represents the channel index in logits, $i$ represents the position index in the map. $T$ is the temperature. $AP_c$ represents the activation map of the $c$-th channel in logits. The effect of activation on each channel is to normalize the image and encode the saliency of scene categories. Figure B.2 shows the activation map $AP^T$ with an index of 1 in logits produced by the teacher model under temperatures $T=1$, 3, and 5. Increasing the temperature can be seen as a smoothing operation on the image. Channel-wise knowledge distillation can guide the student network to produce a similar activation distribution in the foreground saliency [57]. The KL divergence is calculated between the activation maps $AP^T$ and $AP^S$ of the teacher and student model to obtain $L_{soft}$, as shown in Eq. (B.2).

$$L_{soft} = KL(AP^T, AP^S) = \frac{T^2}{C} \sum_{c=1}^{C} \sum_{i=1}^{W \cdot H} AP_{c,i}^T \cdot \log(\frac{AP_{c,i}^T}{AP_{c,i}^S}) \quad (B.2)$$

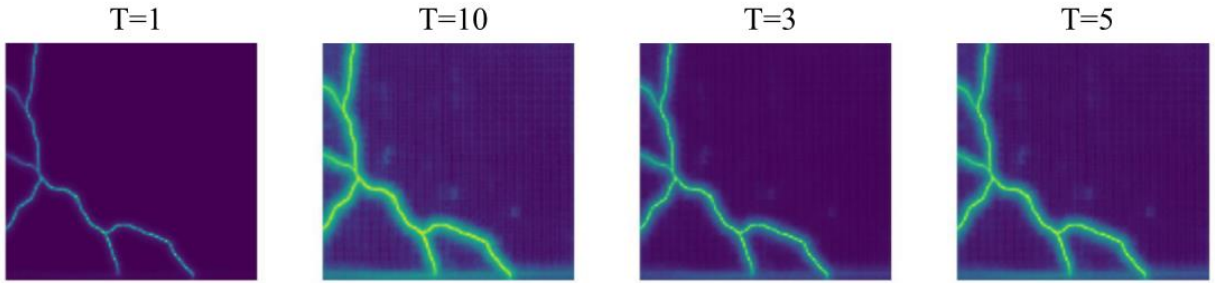

*Figure B.2 The activation map with an index of 1 in logits under different temperature.*

For hard loss, Eq. (8) is used here. The weighted sum of $L_{soft}$ and $L_{hard}$ is the total knowledge distillation loss $L_{kd}$:

$$L_{kd} = L_{hard} + \lambda_s L_{soft} \quad (B.3)$$

where the parameter $\lambda_s$ in Eq. (B.3) is set to 5.0 and temperature $T$ is set to 3.0.

The teacher model is CrackSAM. There are many options for the student model, and ResNet50 and HRNet-W18 used in Section 6 are chosen here. The results of knowledge distillation are shown in Table B.1. Compared to the results in Table 7, both models show improved performance in most cases, demonstrating the effect of knowledge distillation. In some cases, the student model's IOU even exceeds that of the teacher model, which is common because the student model learns both the ground truth and the knowledge from teacher model, to some extent avoiding the tendency to overfit the training set. Taking HRNet as an example, despite the significant capacity gap between the teacher and student model, with a parameter difference of up to 66 times, the performance improvement is still obvious, with an average IOU increase of up to 7.5% in six cases.

*Table B.1 The results of knowledge distillation from different student models.*

| Model | Test set | Noisy test set 1 | Noisy test set 2 | Road420 | Facade390 | Concrete3k |
|---|---|---|---|---|---|---|
| ResNet50 | 0.6441 | 0.5292 | 0.3995 | 0.474 | 0.4357 | 0.6267 |
| HRNet-W18 | 0.6362 | 0.5332 | 0.4134 | 0.5514 | 0.4636 | 0.6565 |

Considering the model performance and parameters, the well trained HRNet-W18 is selected as the student model for edge computing.